# Interpretability methods of machine learning algorithms with applications in breast cancer diagnosis

P. Karatza, K. Dalakleidi, M. Athanasiou, K.S. Nikita, *Fellow, IEEE*

*Abstract*— Early detection of breast cancer is a powerful tool towards decreasing its socioeconomic burden. Although, artificial intelligence (AI) methods have shown remarkable results towards this goal, their "black box" nature hinders their wide adoption in clinical practice. To address the need for AI guided breast cancer diagnosis, interpretability methods can be utilized. In this study, we used AI methods, i.e., Random Forests (RF), Neural Networks (NN) and Ensembles of Neural Networks (ENN), towards this goal and explained and optimized their performance through interpretability techniques, such as the Global Surrogate (GS) method, the Individual Conditional Expectation (ICE) plots and the Shapley values (SV). The Wisconsin Diagnostic Breast Cancer (WDBC) dataset of the open UCI repository was used for the training and evaluation of the AI algorithms. The best performance for breast cancer diagnosis was achieved by the proposed ENN (96.6% accuracy and 0.96 area under the ROC curve), and its predictions were explained by ICE plots, proving that its decisions were compliant with current medical knowledge and can be further utilized to gain new insights in the pathophysiological mechanisms of breast cancer. Feature selection based on features' importance according to the GS model improved the performance of the RF (leading the accuracy from 96.49% to 97.18% and the area under the ROC curve from 0.96 to 0.97) and feature selection based on features' importance according to SV improved the performance of the NN (leading the accuracy from 94.6% to 95.53% and the area under the ROC curve from 0.94 to 0.95). Compared to other approaches on the same dataset, our proposed models demonstrated state of the art performance while being interpretable.

*Index Terms*— Interpretability, Breast Cancer, Ensemble of Neural Networks, Random Forest, Shapley values, Individual Conditional Expectation plot, Global Surrogate model.

## I. Introduction

Breast cancer is the most prevalent cancer in women in the developed and the developing world [1]. Although early detection can decrease the risk of dying from breast cancer, in low and middle - income countries breast cancer diagnosis is performed in very late stages of the disease [2]. To optimize breast cancer survival, early detection remains the best option. Thus, the development of computational models capable of diagnosing breast cancer and enabling the deeper understanding of its pathophysiological mechanisms can pave the way towards decreasing the burden of breast cancer for our society.

Artificial Intelligence (AI) algorithms have been successfully utilized in breast cancer diagnosis. The Wisconsin Diagnostic Breast Cancer (WDBC) dataset, which was used in this study, was used as input in [3] to a Random Forest (RF) algorithm after features selection with Gain ratio, in [4] to a deep learning model which consisted of three hidden layers and in [5] to a k-Nearest Neighbor (k-NN) algorithm after features selection with Neighborhood Component Analysis (NCA).

All the above approaches achieved high levels of diagnostic accuracy, however, due to their internal design, they worked as black boxes, i.e., the underlying reasons that made them form their predictions were unknown. Their black-box nature remains a constraint for widely adopting them in clinical practice, where decisions affect people lives. To overcome the aforementioned limitations of AI methods, interpretability methods have been proposed and used in the recent literature [6], [7]. Interpretability methods can be categorized as intrinsically interpretable models and post hoc (and model-agnostic) interpretable methods [8]. Intrinsically interpretable models are algorithms, such as decision trees, that can be directly understood by humans. For example, by looking the structure of a decision tree, someone can deduce which were the most important input features for the final prediction. Model agnostic methods can explain the predictions of any black-box model (BBM) by training an intrinsically interpretable model with the predictions of the BBM. Other approaches, such as Shapley Values (SV), explain the output of a BBM by means of the features' importance to the final prediction. In the recent literature, to our knowledge, only one study has used inherent interpretable models, such as decision trees, to explain breast cancer diagnosis [9].

In this study, we propose the optimization of breast cancer diagnosis with the use of interpretability methods. The paper aims to: (i) introduce the use of interpretability methods to better understand the underlying mechanisms of breast cancer, (ii) propose a framework for enabling the wide adoption of AI techniques in breast cancer diagnosis by explaining and justifying the prediction of the model to the physician, and (iii) establish a method for optimizing the performance of AI algorithms, such as RF and NN, by feature selection guided by interpretability methods.

## II. Methods

Fig. 1 illustrates the procedure for the creation of the three interpretable machine learning models. First, the AI classifiers are constructed. Then, the interpretability methods are applied to create the final interpretable models.

P. Karatza, K. Dalakleidi, M. Athanasiou and K. S. Nikita are with the Biomedical Simulations and Imaging (BIOSIM) Laboratory, School of Electrical and Computer Engineering, National Technical University of Athens, Greece. (e-mail: gikaratza@gmail.com, knikita@ece.ntua.gr )

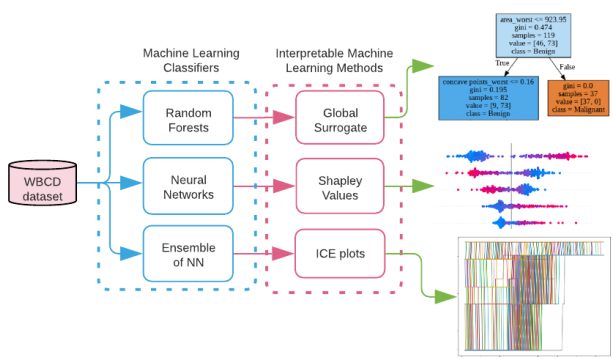

Figure 1. Conceptual framework. A combination scheme based on AI algorithms, interpretable methods, and the final interpretable AI models

*A. Dataset*

In this work, the WDBC dataset [10] which is composed of 569 samples (357 benign, 212 malignant) from breast tumors of patients using Fine Needle Aspirate (FNA) method was used. The FNA samples were digitized in image format. Image processing algorithms deduced the following ten features for each nuclear cell of the sample:

  i. *Perimeter,* the length of the nucleus border.
 ii. *Radius,* the mean value of the radial length of cell nucleus, i.e., the distance from the center of the nucleus to its boundary.
iii. *Area,* approximation of the number of internal pixels in the cell nucleus of the digitized image.
 iv. *Compactness,* computed as $\frac{perimeter^2}{area} - 1$.
  v. *Smoothness,* the difference of the nucleus radius and the average radius of the neighboring nuclei.
 vi. *Concavity,* the number of the severity of concavities in the nuclei border.
vii. *Concave Point,* the number of concave sections of the nuclei border.
viii. *Symmetry,* the differences in the lengths of segments perpendicular to a diametric line of the cell.
 ix. *Fractal,* coast-line approximation measuring the irregularity of the cell nucleus border.
  x. *Texture,* standard deviation of the grayscale values of the nucleus image pixels.

Finally, computation of the mean value, the standard error, and the worst (largest) value of each feature resulted in a vector with thirty features for each sample.

*B. Machine Learning Classifiers*

*Random Forest (RF):* A classifier which consisted of 500 individual decision trees that operate as an ensemble was implemented. Each of the individual decision trees used Gini criterion [11] to measure the quality of a split and make a prediction. Eventually, the class with the highest number of votes became the final prediction of the RF. The RF classifier was chosen due to its efficiency in handling classification problems and its ability to reduce overfitting in decision trees.

*Neural Network (NN):* Due to the complexity of the dataset, the following deep learning model was created. A NN with 1 input, 3 hidden (with 40, 20 and 10 neurons, respectively) and 1 output layer was created. Each layer was fully connected with the previous one and the weights of every neuron was calculated by gradient descent [12]. Batch normalization was adopted to deal with the internal covariance shift [13]. The number of training epochs was 400 and the batch size was 64. The log-sigmoid function was used as the activation function.

*Ensemble of Neural Networks (ENN):* Aiming at further improving performance, an ENN was created. The proposed ENN consisted of NNs that work and make predictions independently, while the result was determined by plurality voting. In our implementation, the three individual neural networks (NN1, NN2, and NN3) of the ENN had the same voting weights. More specifically, individual NNs had the same architecture as the one described earlier with different numbers of hidden layers. NN1 had 2 hidden layers with 25 and 10 neurons, respectively. NN2 had 3 hidden layers with 40, 20 and 10 neurons and NN3 4 hidden layers with 25, 15, 10 and 5 neurons, respectively.

*C. Interpretable Machine Learning Methods*

*Individual Conditional Expectation (ICE):* One way of interpreting a BBM is with an ICE plot that displays curves which describe the effect of changing a specific feature value, while keeping all other features constant, on the final prediction of the model, for each sample of the dataset separately [8]. Thus, ICE provides a detailed description of the model's behavior. One hundred different values belonging to the range between the minimum and maximum of the examined feature were presented in the ICE plots, for each feature separately. Moreover, for visualization purposes, the mean value of the ICE curves was also depicted in a separate plot.

*Global Surrogate (GS):* This method approximates the predictions of a non-interpretable AI method, i.e., of a BBM, with an intrinsically interpretable one [8]. An important advantage provided by the GS method is its flexibility and applicability to any algorithm. In this study, a Decision Tree (DT), an intrinsically interpretable model, was trained to approximate the final prediction of the BBMs. To achieve the best possible performance, the "Gini" criterion was used as the function to measure the quality of each split and nodes were expanded until all leaves contained less than 2 samples. The R-square [8] metric was used to measure how well the GS model replicated the BBMs.

*Shapley Values (SV):* This interpretability method is the only one that has a solid theoretical foundation in game theory. Every feature is used as a player in a game with payout the final prediction of the AI model [8]. Each of the players (features) has the same contribution to the final payout. SV method interprets the effect of every feature to the classification of a sample. In addition, SV show how each feature affects the error of the final prediction from the expected value of the model, i.e. the mean value of the predictions of the model.

### III. RESULTS

*A. Classification*

Table I shows the performance of the proposed AI algorithms in terms of Accuracy (Acc), Sensitivity (Se),

Specificity (Sp) and Area under the Curve (AUC), when evaluated with 10-fold cross validation. The best performance was achieved by the ENN.

TABLE I. PERFORMANCE OF THE PROPOSED ALGORITHMS.

|     | Acc (%) | Se (%) | Sp (%) | AUC |
|-----|---------|--------|--------|-----|
| RF  | 96.49   | 94.37  | 97.75  | 0.96 |
| NN  | 94.1    | 88.36  | 98.74  | 0.93 |
| ENN | 96.6    | 94.94  | 98.48  | 0.96 |

### B. Interpretable Machine Learning Methods

*Individual Conditional Expectation:* The ICE plots for the interpretation of the ENN predictions with respect to the input features "Area Mean" and "Perimeter Mean" are shown in Fig. 2 and 3, respectively. Fig. 2 shows the different variations of the feature "Area Mean" and the corresponding prediction probability values. It can be observed that the probability of a malignant tumor is minimized around a certain value of the feature "Area Mean". This certain value corresponds to a normal shape of the cell nucleus with respect to the radius and perimeter.

The ICE plot of the feature "Perimeter Mean" for the ENN (Fig. 3) shows a similar behavior. As the value of the "Perimeter Mean" increases and the values of the other features remain constant, the shape of the cell nucleus deviates from the normal shape. Thus, the chance of the tumor being malignant increases.

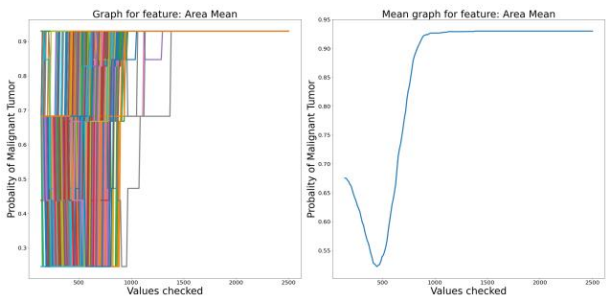

Figure 2. ICE plots of the ENN for the feature "Area Mean"

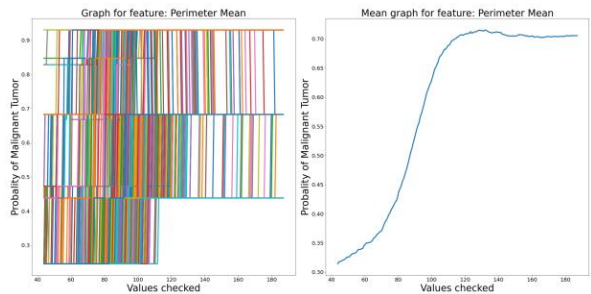

Figure 3. ICE plots of the ENN for the feature "Perimeter Mean"

*Global Surrogate model:* Decision Tree as a GS model achieves the following performance metrics: Acc-88.5%, Se-86%, Sp-90%, Auc-0.88 and R-square 0.49. The performance metrics are high and the R-square metric shows that the interpretable model approximated well the BBM. Fig. 4 shows the interpretation of the predictions of the RF through the Decision Tree as a GS model. The DT separated the samples into classes accurately and their classifications agreed with what is verified by medical evidence [14]. The feature with the greatest influence on the classification of samples was the "Area Worst". Thus, samples whose largest value of the area exceeded the value 927.1 were classified as malignant.

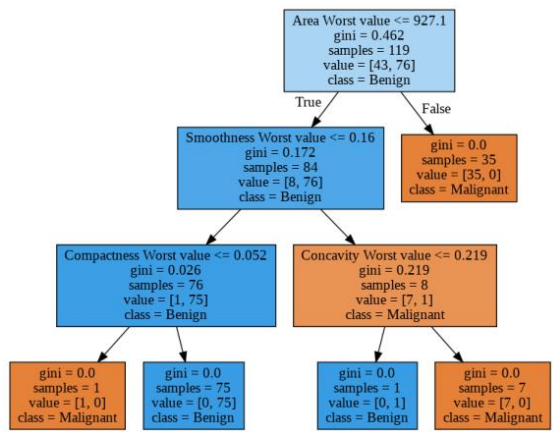

Figure 4. Decision Tree as a Global Surrogate Model for the interpretability of the RF

The ability of the DT to act as a GS model for RF can be observed in Fig. 5, where the predicted values of the RF after Principal Component Analysis (PCA) decomposition are shown on the right and the predicted values of the Decision Tree as a GS model for the RF after PCA decomposition are shown on the left. It can be observed that both the interpretable model and the BBM divided the input space in a similar manner.

After 100 repetitions of the GS method for the RF model, the first 5 most important features were: "Texture Worst", "Fractal Dimension Standard Error", "Area Worst", "Concave Points Mean" and "Smoothness Worst". This subset of features was in complete agreement with medically proven results [14] as cell nuclei with a larger perimeter or larger area are classified as malignant. Also, cells with abnormal borders are more likely to be classified as malignant. Based on the most important features mentioned above, the RF achieved better performance than with all features (Table II).

TABLE II. PERFORMANCE OF THE RF CLASSIFIER AFTER FEATURE SELECTION ACCORDING TO IMPORTANCE WITH RESPECT TO THE GS MODEL.

|                       | Acc (%) | Se (%) | Sp (%) | AUC |
|-----------------------|---------|--------|--------|-----|
| All Features (30)     | 96.49   | 94.37  | 97.75  | 0.96 |
| Selected features (5) | 97.18   | 95.28  | 98.31  | 0.97 |

*Shapley Values:* Fig. 6 shows the summary plot of SV. The summary plot describes the relation between each feature and the final prediction of the model, i.e., the probability that a sample is malignant. More specifically, the position of each point on the y-axis is determined by the feature while its position on the x-axis by the SV. The features are ordered on the y-axis according to their importance in the final prediction. The color of each point determines the value of the feature; higher values are marked in red and lower values are marked in blue. The most important feature, according to Fig.

6, is the "Perimeter Worst". This conclusion can be attributed to the irregular, non-spherical, shape of the cancer cells, which leads to a higher perimeter value.

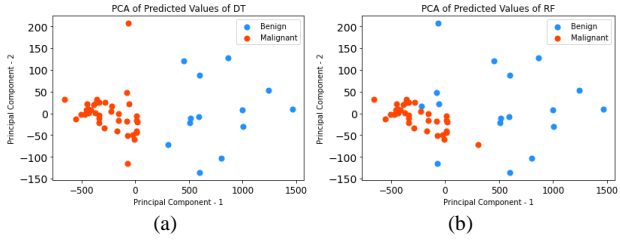

Figure 5. PCA plot: (a) for the predictions of the DT as a GS Model for the RF classifier. (b) for the predictions of the RF classifier.

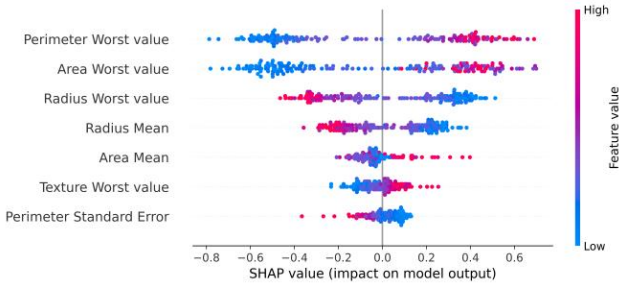

Figure 6. Summary plot of SV for the proposed NN

After 100 repetitions of the SV method, the first 7 most important features were: "Area Worst", "Area Mean", "Area Standard Error", "Perimeter Worst", "Perimeter Mean", "Texture Worst", "Radius Mean". Table III shows the improvement in NN performance following the above feature selection.

TABLE III. PERFORMANCE OF THE NN AFTER FEATURE SELECTION ACCORDING TO THE SV METHOD.

|  | Acc (%) | Se (%) | Sp (%) | AUC |
|---|---|---|---|---|
| All Features (30) | 94.6 | 91.37 | 97.4 | 0.94 |
| Selected features (7) | 95.53 | 93.15 | 97.9 | 0.95 |

### C. Comparison to other approaches

Table IV summarizes the performance metrics achieved by other recent studies carried out on the WDBC dataset as well as our proposed methodology where the selected features from the GS Model were used as input to the RF classifier. It is evident that our method achieves similar performance with the state-of-the-art methods, while being interpretable.

TABLE IV. PERFORMANCE METRICS OF SEVERAL STUDIES IN THE WISCONSIN DIAGNOSTIC BREAST CANCER dataset

| Study | Method | Acc(%) | Se(%) | Sp(%) | AUC |
|---|---|---|---|---|---|
| [3] | RF and Gain Ratio feature selection | 98.77 | 98.8 | 98.7 | 0.99 |
| [4] | Deep Learning model | 99.00 | 98.00 | 98.00 | 0.96 |
| [5] | k-NN and NCA feature selection | 99.00 | 100.00 | 98.00 | 0.95 |
| [9] | Interpretable DT | 96.00 | 100.00 | - | - |
| Ours | Feature Selection based on the GS model and RF classifier | 97.18 | 95.28 | 98.31 | 0.97 |

## IV. CONCLUSION

To address the need for AI guided breast cancer early diagnosis, in this study, we investigated the use of various AI methods, including Random Forests, Neural Networks, and Ensembles of Neural Networks. Different interpretability techniques, namely the Global Surrogate model, the Individual Conditional Expectation plots, and Shapley values, were investigated towards the explanation of the models' predictions. The WDBC dataset was used for training and evaluation purposes. The developed models achieved satisfying discrimination performance, while the application of interpretability methods ensured that their predictions were not arbitrarily correct and could be verified by medical knowledge. Feature selection based on the obtained explanations was introduced towards optimizing the models' performance. Future work will focus on applying combinatorial interpretability approaches as well as using advanced deep learning models, such as one-dimensional Convolutional Neural Networks, towards the development of interpretable, highly expressive AI models for breast cancer diagnosis.